\documentclass[sigconf, nonacm, review=false]{acmart}

\usepackage{multirow, dsfont, enumitem}

\AtBeginDocument{%
  \providecommand\BibTeX{{%
    \normalfont B\kern-0.5em{\scshape i\kern-0.25em b}\kern-0.8em\TeX}}}
    
\usepackage{graphicx}
\usepackage{subfig}
\usepackage[percent]{overpic}
\usepackage[countmax]{subfloat}
\usepackage{stfloats}
\usepackage{mathtools}
\usepackage{hyperref}
\usepackage{bbm}
\usepackage{geometry}
\usepackage{tabularx}
\usepackage{booktabs}

\graphicspath{ {./images/} }

\begin{document}

%To be submitted to https://fashionxrecsys.github.io/fashionxrecsys-2023/

% efficient learning, efficient architectures, evaluation

\title{Efficient Large-Scale Visual Representation Learning And Evaluation}

\author{Eden Dolev, Alaa Awad, Denisa Roberts}
\authornote{Denisa Roberts is the corresponding author.}
\author{Zahra Ebrahimzadeh, Marcin Mejran, Vaibhav Malpani, Mahir Yavuz}
\affiliation{%
  \institution{Etsy}
  \streetaddress{117 Adams St}
  \city{Brooklyn}
  \state{NY}
  \country{USA}
  \postcode{11201}
}
\email{{edolev, aawad, denisaroberts}@etsy.com}

% \author{Eden Dolev}
% \affiliation{%
%   \institution{Etsy}
%   \streetaddress{117 Adams St}
%   \city{Brooklyn}
%   \state{NY}
%   \country{USA}
% }
% \email{edolev@etsy.com}

% \author{Alaa Awad}
% \affiliation{%
%   \institution{Etsy}
%   \streetaddress{117 Adams St}
%   \city{Brooklyn}
%   \state{NY}
%   \country{USA}
%   \postcode{11201}
% }
% \email{aawad@etsy.com}

% \author{Denisa Roberts}
% \authornote{Corresponding author.}
% \affiliation{%
%   \institution{Etsy}
%   \streetaddress{117 Adams St}
%   \city{Brooklyn}
%   \state{NY}
%   \country{USA}
% }
% \email{denisaroberts@etsy.com}

\renewcommand{\shortauthors}{Dolev, Awad, and Roberts}

\begin{abstract}

Efficiently learning visual representations of items is vital for large-scale fashion recommendations in e-commerce. In this article we compare several pretrained efficient backbone architectures, both in the convolutional neural network (CNN) and in the vision transformer (ViT) family. We describe challenges in e-commerce vision applications at scale and highlight methods to efficiently train, evaluate, and serve visual representations. We present ablation studies evaluating visual representations in several downstream tasks. To this end, we present a novel multilingual text-to-image generative offline evaluation method for visually similar fashion recommendation systems. Finally, we include online results from deployed machine learning systems in production on a large scale e-commerce platform.

\end{abstract}




\maketitle

\begin{figure}[!tbp]
    \centering
    \subfloat[\centering Query image]{{\includegraphics[scale=0.3]{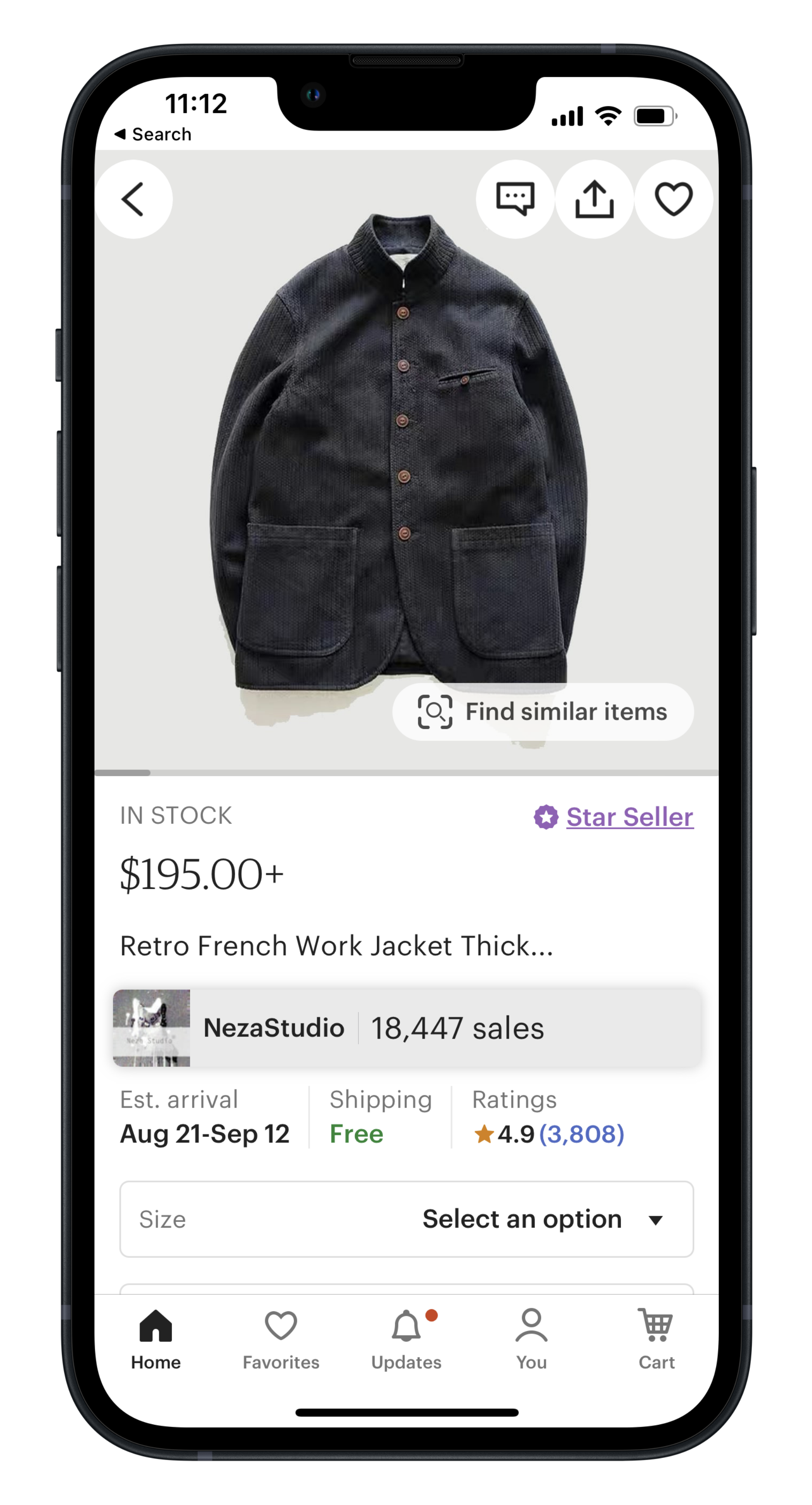}\label{fig:vis_sim_query} }}%
    \subfloat[\centering Find visually similar result images]{{\includegraphics[scale=0.3]{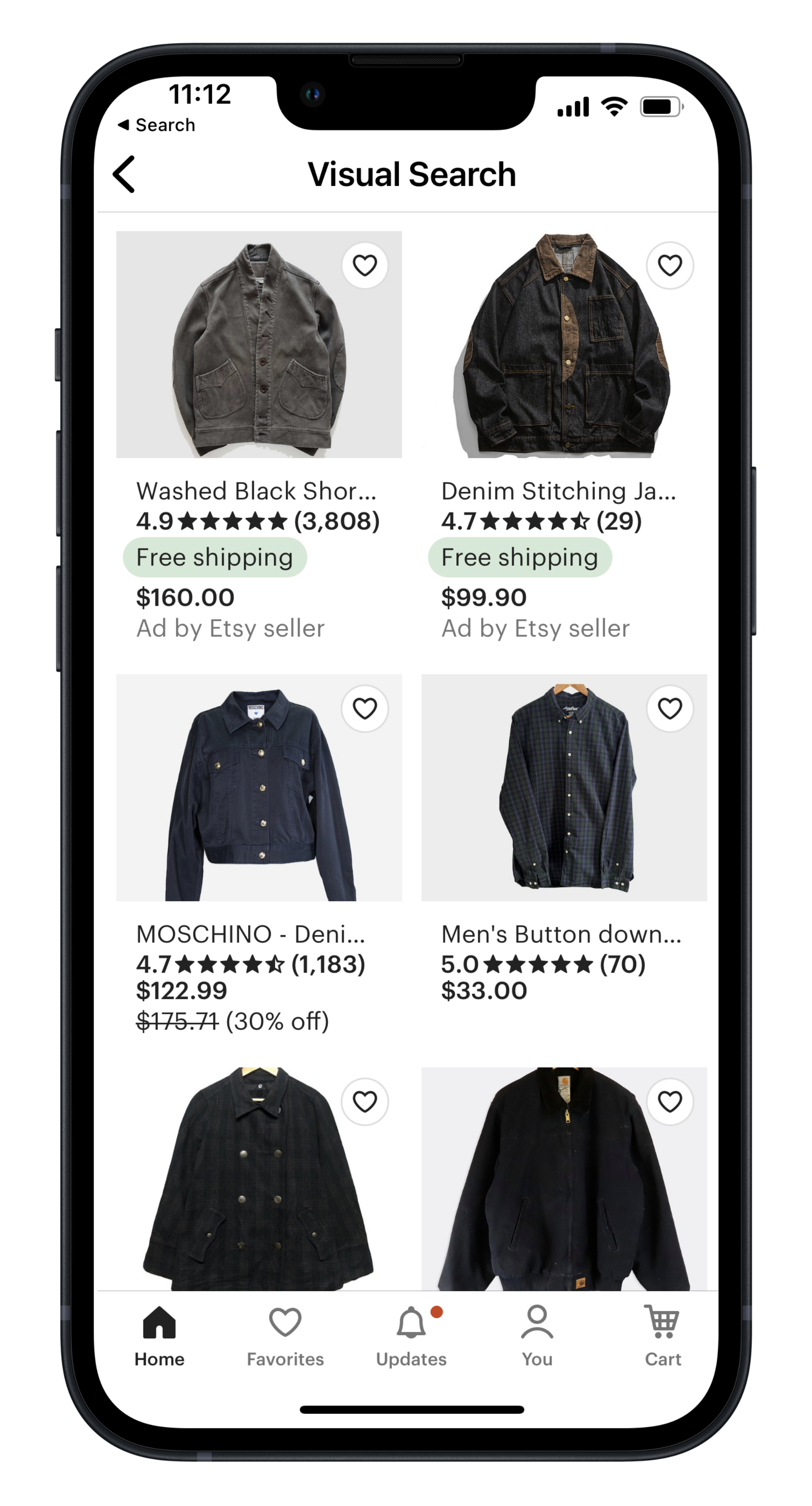}\label{fig:vis_sim_results}}}%
    \caption{Visual representations power visual search and visually similar recommendations. A query image is inferred online and the top-k nearest neighbor recommendations are retrieved.}%
    \label{fig:visual_apps}%
\end{figure}

\section{Introduction}

Large-scale visual representation learning has emerged as a powerful technique for addressing a wide range of computer vision tasks. For e-commerce in particular, techniques have been applied across item recognition, visual search, content moderation, recommendations, and advertising \cite{jing2015visual, yang2017visual, du2022amazon, zhang2018visual}. Online fashion is inherently a computer vision application, especially on e-commerce platforms which curate unique, hand-made, and vintage items of high visual diversity. It remains a challenge to efficiently retrieve visually similar items from large-scale, diverse collections of images. User queries of images tend to include photos from real-life or uploaded images taken from another source. However, seller uploaded images tend to have higher quality, better lighting, clear backgrounds, and face-forward angles. The heterogeneous sources create a semantic and visual gap when comparing visual similarity and high quality visual representations are necessary. Expressive visual representations are learned through complex and costly deep learning architectures, memory-hungry high-dimensional vectors, and carefully crafted training regimens. Furthermore, in search-by-text retrieval there can be a gap between a text query and the visual imagination for the sought item, especially in visually intensive applications, such as retrieval and ranking of unique, custom, and hand-crafted fashion or interior design items. Advances in text-to-image generative AI can help bridge the gap and the impact of large scale visual models stands to expand across recommender and information retrieval domains. Offline evaluation is of vital importance for effectively learning visual representations and remains a challenge. Qualitatively and quantitatively evaluating visual representations can be subjective and often a ground-truth dataset is not readily available. Visual representations in e-commerce are also evaluated through the benefit they bring to end users, measured by ad clicks or item purchases. We evaluate this benefit offline and use it to guide the representation learning process. Visual representation learning presents infrastructure challenges as well. Applications on mobile devices are latency sensitive and efficient inference is paramount. Furthermore, learning and deploying high quality visual representations at scale can rapidly exceed departmental budgets on cloud costs and energy consumption. Efficiency enables increased adoption.

This article presents methods to tackle these challenges effectively by efficiently designing, learning, and evaluating visual representations. To mitigate for heterogeneity of images and for text-query-to-user-stylistic-intent gap in fashion recommendations we introduce a novel multilingual text-to-image generative method for visually similar items retrieval and recommendation evaluation.

\section{Related Work}

\begin{figure*}[!tbp]
    \centering
    \subfloat{{\includegraphics[scale=0.5]{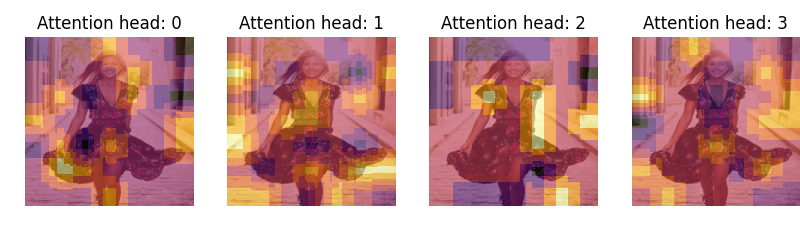} }}%
     \hspace{0.1cm}
    \subfloat{{\includegraphics[scale=0.5]{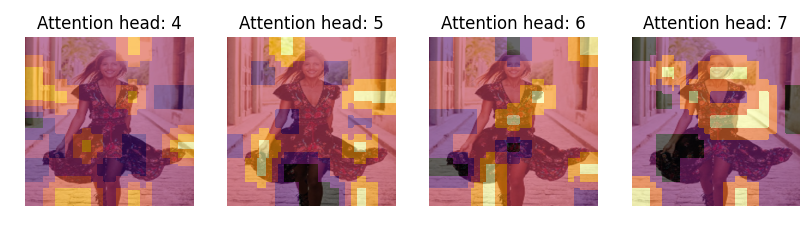} }}%
    \caption{Probing the EfficientFormer l3 pretrained representations through attention heatmaps. Image of a woman wearing a boho chic dress walking on a narrow street between tall buildings.}%
    \label{fig:attn}%
\end{figure*}

Visual representations are used across visually similar recommendations, visual search \cite{du2022amazon, jing2015visual, zhai2017visual}, and stylistically complementary fashion recommendations \cite{kang2019complete, li2020bootstrapping}. However, learning expressive representations \cite{bengio2013representation} and leveraging them at scale is a costly technical challenge. We can think about efficiency in deep learning along three axes \cite{menghani2023efficient}: efficiency in architecture design, efficiency in training, and efficiency in inference and serving. \textbf{Architecture design efficiency.} Convolutional Neural Networks (CNN) first employed for visual representations \cite{he2016deep, sandler2018mobilenetv2, tan2019efficientnet} captured local spatial patterns in the image, such as edges and corners. More recently, vision transformers (ViT) \cite{dosovitskiy2020image} emerged as backbones. Although ViT lack the spatial inductive biases of CNN, they outperform CNN when trained with a large enough amount of data and may be more robust to domain shifts \cite{bhojanapalli2021understanding}. However ViT can be prohibitively large and expensive to train and serve and more efficient ViT have emerged \cite{liu2021swin, dong2022cswin, yu2022metaformer, vasu2023fastvit}. Architectural efficiency is typically tackled through data efficiency \cite{touvron2021training}, clever pooling \cite{yu2022metaformer}, layer dropping, efficient normalization \cite{li2022efficientformer, li2022rethinking}, efficient attention \cite{bolya2022hydra, ryali2023hiera, shaker2023swiftformer}, or hybrid CNN-transformer designs \cite{mehta2021mobilevit, vasu2023fastvit}. \textbf{Training efficiency.} Learning efficiency can be achieved through label efficiency, transfer learning \cite{luo2017label, hoffman2013efficient, li2021efficient}, pretraining through self-supervision \cite{he2022masked}, or efficient finetuning with compression \cite{hu2021lora, roberts2020qr}. Multitask learning \cite{caruana1997multitask} is an efficiency inducer as demonstrated across machine learning problems \cite{beal2022billion, zhai2019learning, bell2020groknet, roberts2019neural, radford2019language, baltescu2022itemsage}. Efficient training is further enhanced by effective offline evaluation methods. \textbf{Inference and serving efficiency.} Visual representations are included in downstream tasks such as retrieval, ranking, and recommendations. Efficient top-k retrieval is typically achieved through approximate nearest neighbors \cite{johnson2019billion}. In downstream rankers, pretrained visual representations can be served with the model or from feature stores \cite{awad2023adsformers, baltescu2022itemsage, beal2022billion}. In tasks such as visual search efficient online inference must be achieved through efficient architectures as well as efficient serving infrastructure \cite{li2022efficientformer, awad2023adsformers, li2023snapfusion}. 

\section{Methodology}

We compare several readily available pretrained backbones which we finetune under a few different settings and evaluate comprehensively offline and online. We focus on finetuning only, due to publicly available pretrained weights, an efficiency inducer.

\subsection{Efficient Architectures in Visual Representation Learning}
\label{efficientnetbackbone}

The EfficientNet family of models uniformly scale network width, depth, and resolution using a set of fixed scaling coefficients. To upscale we can increase the image size by $\gamma^N$, network depth by $\alpha^N$, and width by $\beta^N$, where $\gamma, \alpha, \beta$ are determined by grid search. We employ EfficientNetB0, the smallest size model in the EfficientNet family. We experiment with the Base ViT16 as a standard vanilla transformer baseline. Figure 1 in \cite{li2022efficientformer} compares accuracy versus latency tradeoffs for a few vision transformer architectures and we experiment with the EfficientFormer l1 and l3, the data-efficient but high-latency DeiT large \cite{touvron2021training}, and MobileVit \cite{mehta2021mobilevit}. The EfficientFormerl3 scores 2.7ms for iPhone12 inference with 30mln parameters, versus EfficientNetB0 at 1.7ms with 5mln parameters \cite{li2022efficientformer}. The EfficientFormer achieves efficiency through multiple blocks downsampling and only employing attention at the last stage. We visualize attention maps of the EfficientFormer l3 pretrained backbone in Figure \ref{fig:attn}. Similarly to other vision transformer architectures \cite{raghu2021vision} we see how different heads learn to focus on different salient parts of the image. To extract the image embedding in CNN models, we replace the final convolutional layer in the backbone and attach a smaller 256-dimensional CNN layer for memory efficiency, followed by a batch normalization layer, a swish activation, and a global average pooling layer which aggregates the convolutional output into the final representation. Then we attach $m$ classification heads, where $m$ is the number of tasks. From Equation \ref{eq:one}, if $X_j$ is the output of the $j^{th}$ layer of the pretrained backbone,

\begin{equation}
X_{shared} = GlobalAvgPool(Swish(BN(Conv_{d}(X_j)))),
\label{eq:one}
\end{equation}

\noindent then $y_m = Linear_{m}(X_{shared})$, where $y$ is the prediction from each of the $m$ tasks. We extract ViT's last hidden state projected down by a dense layer with a $tanh$ activation and layer normalization. Then classification heads are added as for the CNN with batch normalization being swapped for layer normalization. When extracting a 512d representation from the EfficientFormer, we average pool the last hidden state and skip the down projection layers.

\begin{figure*}[!tbp]
\includegraphics[scale=0.35]{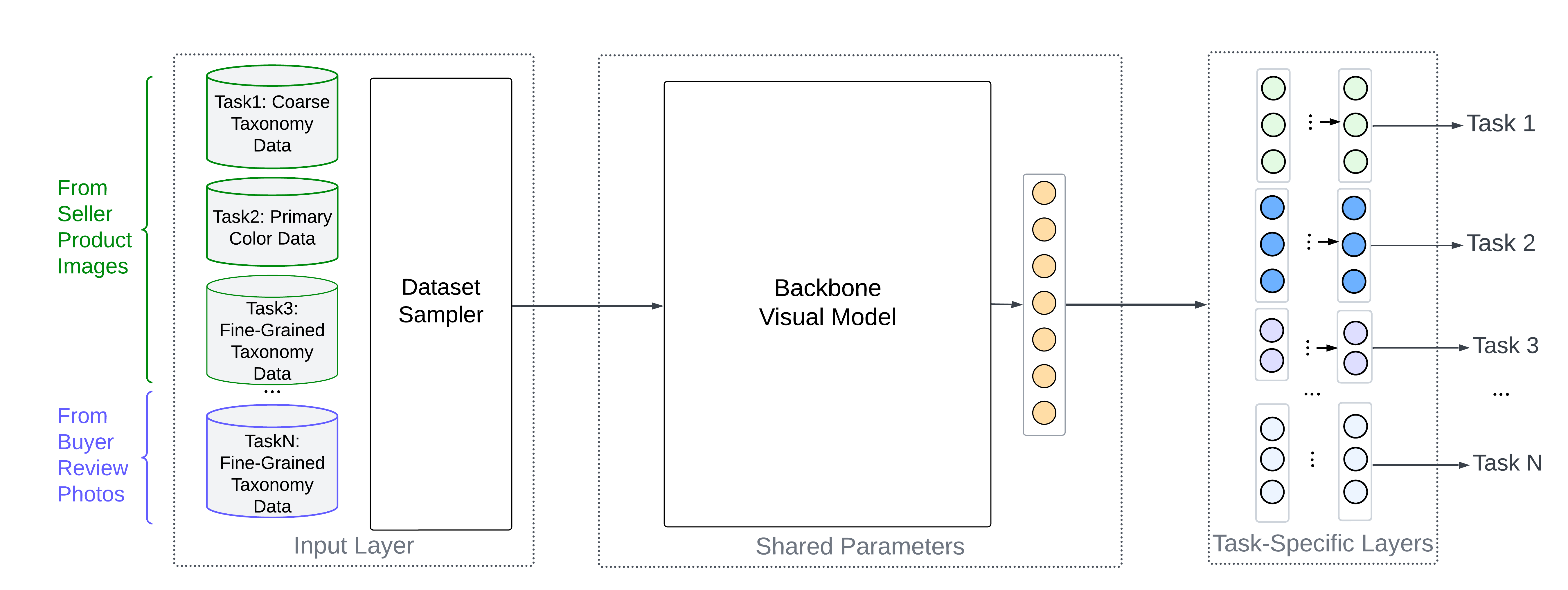}
\centering
\caption{Multitask visual representations training architecture. The data sampler combines examples from an arbitrary number of datasets corresponding to respective classification heads. The visual representation layer is shared before the heads.}
\label{fig:viz-multitask-arch-3ds}
\end{figure*}

\subsection{Efficient Training} 

\subsubsection{Deep Metric Learning}
The predominant approach for learning visual representations that capture similarity is deep metric learning. We employ a three-instance feed-forward network with shared parameters. A positive, negative, and anchor image are fed to the network and triplet loss \cite{hoffer2015deep} minimizes the distance between intra-class image representations and maximize the distance between inter-class image representations. The training dataset includes 10 images for each item. Two images for the same item are considered positive, while an image from any other item is considered negative. A major challenge with deep metric learning is sampling informative triplets. Sampling negatives that are too easy results in slow convergence due to near-zero loss while negatives that are too hard can destabilize training. While larger batch sizes help mining useful negatives, in practice we are limited by hardware memory constraints. Multitask classification can mitigate these challenges.

\subsubsection{Multitask Learning}

Representations learned using common attributes as multiple supervision signals encode commonalities efficiently to perform well in diverse downstream tasks. A representation learned in single-task classification to the item's taxonomy does not capture other visual attributes-colors, shapes, materials. We employ four classification tasks: a top-level taxonomy task with 15 top level categories of the taxonomy tree as labels; a fine-grained taxonomy task, with 1000 fine-grained leaf node item categories as labels; a primary color task; a fine-grained taxonomy task (review photos), where each example is a buyer-uploaded review photo of a purchased item with 100 labels sampled from fine-grained leaf node item categories.

Structured metadata input by sellers, such as colors, materials, and shapes, tend to be sparse. For this reason we implemented a framework that reads from an arbitrary set of disjoint datasets as seen in Figure \ref{fig:viz-multitask-arch-3ds}. A data sampler interleaves a uniform number of examples from each dataset into each training minibatch. Each minibatch example has one real label. All other labels take a special sentinel value, which is ignored in the loss calculation. A layer rescales, resizes, crops, rotates, and jitters the images at train time. For CNNs, we train new layers from scratch for one epoch with the backbone layers frozen, to avoid excessive computation and overfitting. We then unfreeze 75 layers from the top of the backbone and finetune for nine additional epochs, for efficient learning. We make a few efficiency inducing tradeoffs in training ViT. We finetune with a 224X224 resolution since longer sequences from the recommended 384X384 \cite{dosovitskiy2020image} lead to larger training budgets. Most of the ViT output a down-projected 256d representation, finetuned together with the backbone for a constant number of epochs.

\begin{figure*}[!tbp]
    \centering
    \subfloat{\includegraphics[scale=0.2]{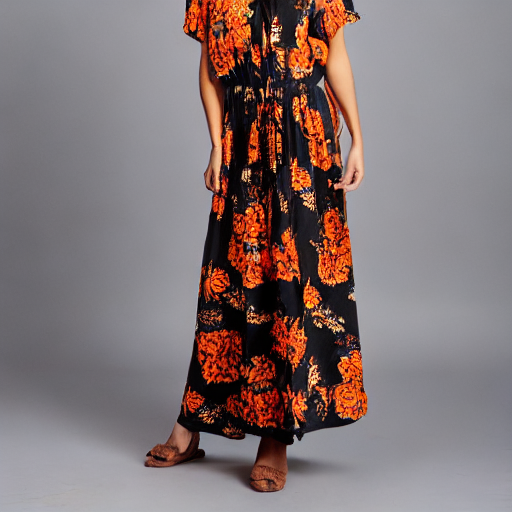}\includegraphics[scale=0.2]{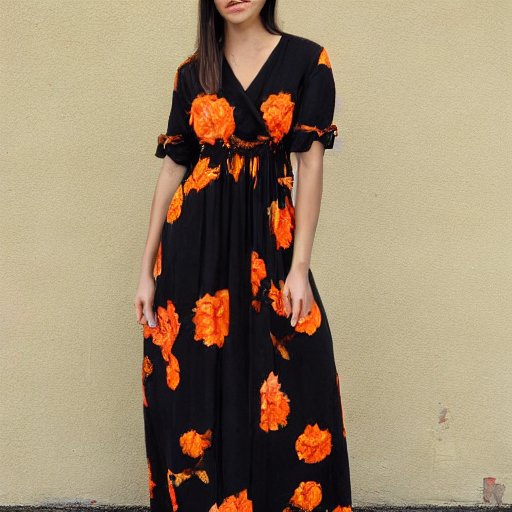} \includegraphics[scale=0.2]{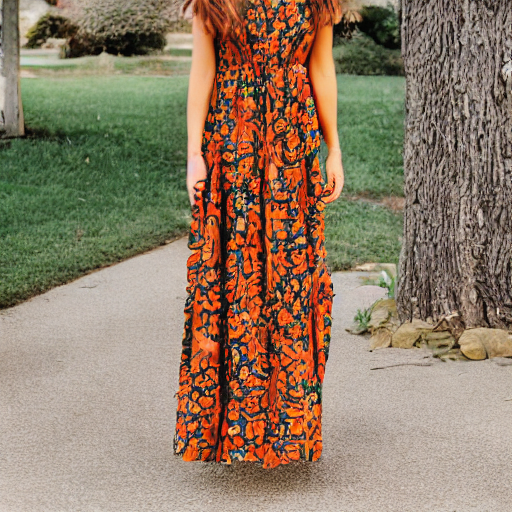} \includegraphics[scale=0.2]{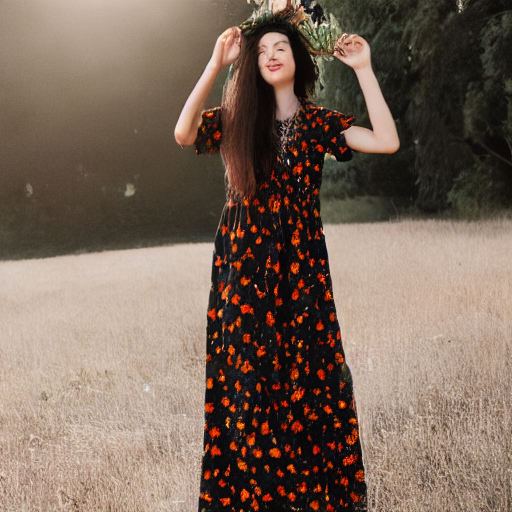} \includegraphics[scale=0.2]{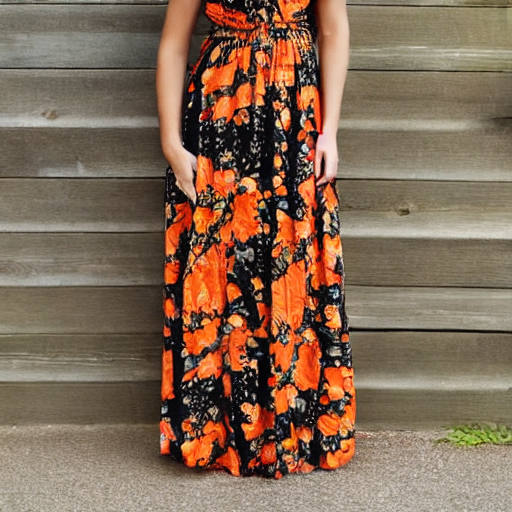}}%
    \captionsetup{labelformat=empty}
    \caption{Random sample of $n$ generated image queries for the English text query \textit{``Black bohemian maxi dress with orange floral pattern.''}}%
    \subfloat{\includegraphics[scale=0.2]{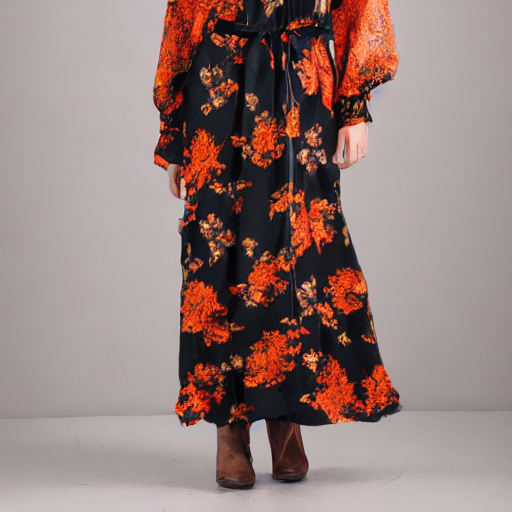}\includegraphics[scale=0.2]{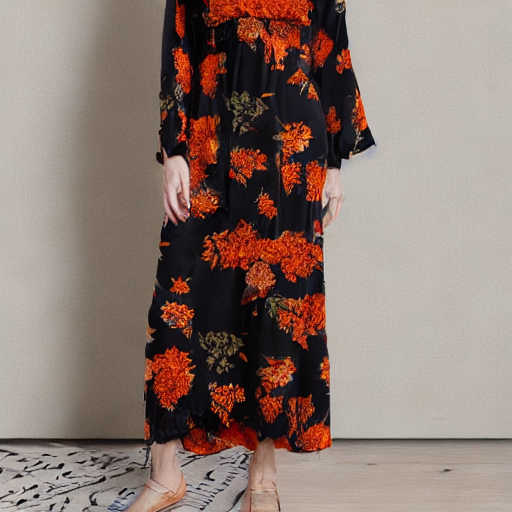} \includegraphics[scale=0.2]{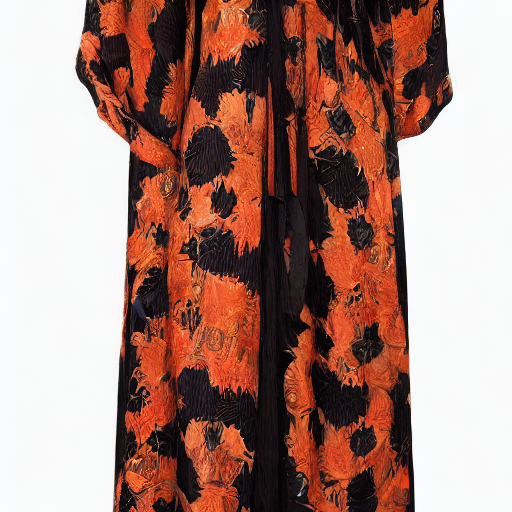} \includegraphics[scale=0.2]{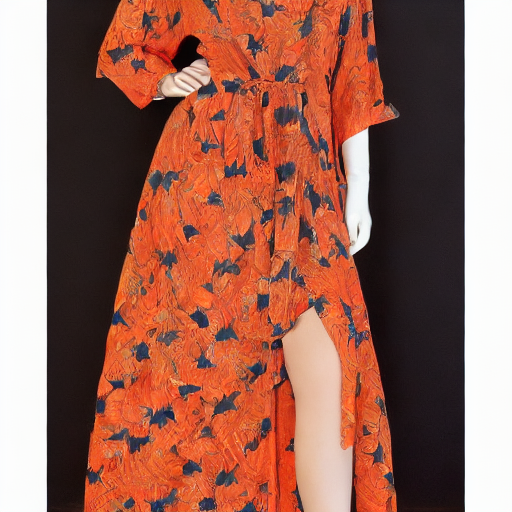} \includegraphics[scale=0.2]{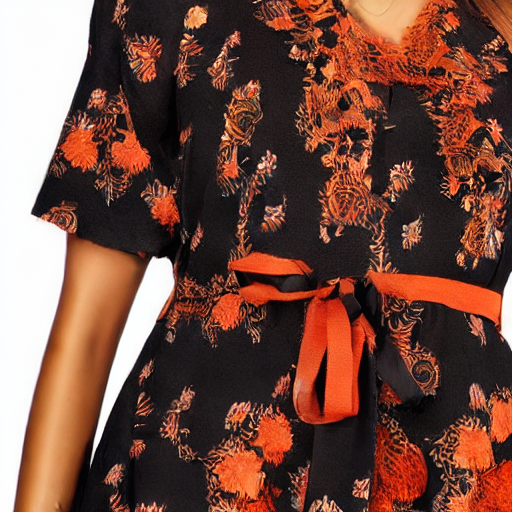}}%
     \caption{Random sample of $n$ generated image queries for the French text query \textit{``Robe longue noire boheme a motif fleur orange''}.}%
     \captionsetup{labelformat=simple}
\caption{Text-to-image generation}
\label{fig:gen}%
\end{figure*}

\subsection{Evaluating Visual Representations for Efficient Learning}

 Next we present offline evaluation methods and heterogeneous datasets which can be employed to guide the finetuning of visual representation models toward accuracy on specific downstream tasks, such as visually similar recommendations.

\subsection{Offline Retrieval Evaluation Methods}
\label{sec:retrieval_eval}
We design an evaluation method to track model performance during and after training on four nearest neighbor retrieval tasks. Each evaluation task has two associated datasets: ``queries'' and ``candidates''. The ``candidates'' dataset feeds a brute force nearest neighbor index, and the ``queries'' dataset is used to look up the index. The index is constructed on the fly after each epoch to accommodate for embeddings changing between training epochs. Each lookup yields $K$ nearest neighbors. A Keras callback is invoked to compute a Recall@K metric for $K\in \{5, 10\}$ for each task. The ``row-wise recall'' over $N$ rows is $R@K = \frac{TP_{K}}{N}$, where the number of true positives, $TP$ is defined in Equation \ref{eq:two} as 

\begin{equation}
TP_{K} = \sum_{i=1}^{N} \Bigg(\mathbbm{1}{[y_{i,j\in{K}} = y_{i, \text{query}}]}\Bigg).
\label{eq:two}
\end{equation}

Each row that yields at least one true positive in its $K$ candidates is considered a positive example. We then divide the number of true positives by the number of rows to obtain the recall. We can use these Recall@K metrics to efficiently early stop the neural network training. The ``Intra-Item'' dataset contains groups of seller-uploaded images of the same item. The query and candidate examples are randomly selected seller-uploaded images of an item. In the ``Intra-Item with Reviews'' set the query image is a randomly selected buyer-uploaded review image of an item and seller-uploaded images are candidate examples. For both datasets, a candidate image is considered a positive example if it is associated with the same item as the query image. ``Visually Similar Ad Clicks'' associates seller-uploaded images of items seen on the visually similar surface on mobile with those that have been subsequently clicked. A candidate image is considered a positive example if its associated item has been clicked by a user when viewing the query image. Each evaluation dataset contains 15,000 records for building the index and 5,000 query images to look up the index.

\subsection{Offline Evaluation Method using Efficient Text-to-Image Diffusion Models}

Language guided counterfactual test images are used in \cite{prabhu2023lance} to stress test image classification models. Building on this concept we propose a novel method for visual representations evaluation using text-to-image diffusion models. From ample historical multilingual text query logs we build a retrieval dataset that bridges the semantic gap between text-based queries and clicked image candidates. A diffusion model \cite{rombach2022high, saharia2022photorealistic} generates high-quality images which become image queries. The ``candidates'' are images from clicked items corresponding to the source text query in the logs. The source-candidate image pairs form the new evaluation dataset which is then used within the retrieval callbacks. Any given user typing a text query has a different imagination of what that garment might look like. For this reason we choose to generate several images for each query, formally considering a prior $p(\sigma)$, where $\sigma$ is the personal fashion style parameter, $\sigma \in S$, and $S$ is the fashion style space. We generate a random sample of length $n$ from the conditional distribution $p(\mathbf{x}, \sigma|q)$ over all possible images $\mathbf{x} \in X$ that can be generated from the seed text query $q$, for all fashion styles $\sigma \in S$. Then, by Bayes' Rule, the selected image queries are generated by the conditional process $p(\mathbf{x}|q, \sigma) \propto \frac{p(\mathbf{x}, \sigma|q)}{p(\sigma|q)}$. We employ the uniform distribution, but other probability distributions can be used. We randomly select one of the $n$ generated images to replace the text query with an image query.

Large memory requirements and high inference latency are challenges in employing text-to-image generative models at scale. Efficient text-to-image generation is an area of active research \cite{li2023snapfusion, bolya2023token, liu2023cones}. We employ an open source \cite{von-platen-etal-2022-diffusers} fast stable diffusion model through token merging \cite{bolya2023token}. The method provides a $50\%$ speedup of inference, with five times memory consumption reduction, though results depend on the underlying patched model. The query dataset starts with 5,000 text queries and the index dataset has 41,000 examples, with several clicked candidates for each query. A qualitative assessment in Figure \ref{fig:gen} shows a generated sample for the English text query \textit{``Black bohemian maxi dress with orange floral pattern''}. The generated images include pleasant variations. Interestingly, the fashion models' facial details can suffer some loss while garment patterns remain clear. The diffusion model is multilingual like other transformers in IR \cite{olteanu2021multilingual, lavi2021learning}. By generating via the French translation of the text query, \textit{``Robe longue noire boheme a motif fleur orange''}, the images display a somewhat French style.

\begin{table*}
    \centering
    \tiny
    \caption{Classification accuracy metrics. MT1 stands for multitask learning with three tasks, while MT2 has four tasks; ST is single-task classification.}
    \label{tab:accmetrics}
    \begin{tabularx}{\linewidth}{Xlrrrrrrrr}
        \toprule
        Training Architecture & Model Architecture & \multicolumn{2}{c}{Top-Level Taxonomy} & \multicolumn{2}{c}{Fine-Grained Taxonomy} & \multicolumn{2}{c}{Primary Color} & \multicolumn{2}{c}{Fine-Grained Taxonomy (Reviews)}\\ 
                       &             & Top1-Acc & Top5-Acc &  Top1-Acc & Top5-Acc   & Top1-Acc & Top5-Acc &  Top1-Acc & Top5-Acc           \\ 
        \midrule
        ST Linear Probe & EfficientNetB0 Backbone & - & - & 0.2035 & 0.4065 & - & - & - & - \\
        MT2 Linear Probe & EfficientNetB0 Backbone & 0.5084 & 0.8583 & 0.2000 & 0.4171 & 0.3681 & 0.7990 & 0.1581 & 0.3501 \\
        Single-task & EfficientNetB0 Backbone & - & - & 0.2883 & 0.5230 & - & - & - & - \\
        MT1 & EfficientNetB0 Backbone & 0.5980 & 0.8947 & 0.2824 & 0.5174 & 0.4890 & 0.8576 & - & - \\
        MT2 & EfficientNetB0 Backbone & 0.6206 & 0.8992 & 0.2905 & 0.5292 & 0.5040 & 0.8618 & 0.2137 & 0.4305 \\
        Pseudo Zero-Shot & ViT Base16 Backbone & - & - & 0.0007 & 0.0052 & - & - & - & -  \\
        ST & ViT Base16 Backbone & - & - & 0.3606 & 0.6239 & - & - & - & -  \\
        MT1 & ViT Base16 Backbone & 0.6760 & 0.9250 & 0.3560 & 0.6120 & \textbf{0.5800} & 0.8900 & - & -  \\
        MT2 & ViT Base16 Backbone & \textbf{0.6800} & \textbf{0.9300} & \textbf{0.3700 }&\textbf{ 0.6300} & 0.5700 & \textbf{0.8900} & \textbf{0.2700} &  \textbf{0.5300} \\
        ST & DeiT Backbone & - & - & 0.2590 & 0.5187 & - & - & - & -  \\
        MT2 & DeiT Backbone & 0.6693 & 0.9215 & 0.3441 & 0.6006 & 0.5657 & 0.8866 & 0.2603 & 0.5048  \\
        ST & MobileVit xSmall Backbone & - & - & 0.2303 & 0.4594 & - & - & - & -  \\
        MT2 & MobileVit xSmall Backbone & 0.6023 & 0.8955 & 0.2577 & 0.4912 & 0.5242 & 0.8752 & 0.1876 & 0.3942  \\
        MT2 & EfficientFormerl1 Backbone 256 & 0.6203 & 0.9006 & 0.2981 & 0.5305 & 0.523 & 0.8682 & 0.2175 & 0.4398  \\
        MT2 & EfficientFormerl1 Backbone 448 & 0.6203 & 0.9000 & 0.2981 & 0.5301 & 0.5230 & 0.8882 & 0.2175 & 0.4398 \\
        ST & EfficientFormerl3 Backbone 256 & - & - & 0.3018 & 0.5298 & - & -  \\
        ST & EfficientFormerl3 Backbone 512 & - & - & 0.3076 & 0.5561 & - & - & - & - \\
        MT2 & EfficientFormerl3 Backbone 256 & 0.6345 & 0.9037 & 0.3128 & 0.5458 & 0.5169 & 0.8600 & 0.2332 & 0.4583  \\
        MT2 & EfficientFormerl3 Backbone 512 & 0.6425 & 0.9080 & 0.3079 & 0.5457 & 0.5509 & 0.8804 & 0.2269 & 0.4477  \\
        \bottomrule
    \end{tabularx}
\end{table*}

\section{Experiments}

\subsection{Ablation Studies}

We perform ablation studies to evaluate backbone choices and training paradigms. Only the EfficientNetB0 is finetuned in the triplet learning paradigm. All the ViT pretrained weights, publicly available from \cite{wolf2019huggingface}, are finetuned in single-task (ST) classification or multitask with three (MT1) or four tasks (MT2). Additionally, we compare the performance of a linear probe, where we only train the added classification head and keep pretrained weights frozen for the CNN. Projecting down the ViT representation to 256d poses a challenge for evaluation of a vision transformer in the zero-shot (no finetuning) paradigm. Thus we employ a pseudo zero-shot method which finetunes all layers on eight examples to lightly update the down-projection layer. Classification accuracy metrics are in Table \ref{tab:accmetrics} and retrieval callback metrics are in Table \ref{tab:all_callbacks}. All results are for the 256d representation in all tables, unless otherwise noted. The multitask finetuned ViT Base16 outperforms the other architectures in classification accuracy, but it comes at the cost of high latency, which makes it unusable in latency sensitive applications where the query image is inferred online.

\begin{table*}[bhtp]
    \centering
    \tiny
    \caption{Recall$@$5 and Recall$@$10 metrics for three different similarity tasks. MT1 stands for multitask classification with three tasks, while MT2 has four tasks; ST is single-task classification.}
    \label{tab:all_callbacks}
    \begin{tabularx}{\linewidth}{XXrrrrrrrr}
        \toprule
        Training Architecture & Model Architecture & \multicolumn{2}{c}{Item-Item Retrieval} & \multicolumn{2}{c}{Review-Item Retrieval} & \multicolumn{2}{c}{Visually-Similar Clicks} \\ 
                  &     &          R@5 & R@10        &      R@5 & R@10            &            R@5 & R@10           \\ 
        \midrule
        Triplet & Pretrained EfficientNetB0 Backbone & 0.696279 & 0.719459 & 0.220375 & 0.256625 & 0.656460 & 0.724788 \\
        Triplet & Finetuned EfficientNetB0 Backbone & \textbf{0.8421} & \textbf{0.8592} & 0.3539 & 0.4069 & 0.6875 & 0.7562\\
        Zero-Shot & EfficientNetB0 Backbone & 0.7226 & 0.7527 & 0.2781 & 0.3174 & 0.7150 & 0.7839 \\
        ST Linear Probe & EfficientNetB0 Backbone & 0.7258 & 0.7565 & 0.2870 & 0.3335 & 0.7375 & 0.8104 \\
        MT2 Linear Probe & EfficientNetB0 Backbone & 0.7335 & 0.7661 & 0.3039 & 0.3550 & 0.7526 & 0.8260 \\
        ST & EfficientNetB0 Backbone & 0.7272 & 0.7619 & 0.2781 & 0.3321 & 0.7210 & 0.7956 \\
        MT1 & EfficientNetB0 Backbone & 0.7573 & 0.7890 & 0.3021 & 0.3545 & 0.7439 & 0.8182 \\
        MT2 & EfficientNetB0 Backbone & 0.7674 & 0.7994 & 0.3455 & 0.3990 & 0.7691 & \textbf{0.8456} \\
        Pseudo Zero-Shot & ViT Base16 Backbone & 0.6903 & 0.7208 & 0.2506 & 0.2989 & 0.5883 & 0.6577  \\
        ST& ViT Base16 Backbone & 0.7035 & 0.7378 & 0.2925 & 0.3493 & 0.6467 & 0.7223  \\
        MT1 & ViT Base16 Backbone & 0.7435 & 0.7758 & 0.3140 & 0.3745 & 0.6311 & 0.7056  \\
        MT2 & ViT Base16 Backbone & 0.7634 & 0.7938 & 0.3598 & 0.424 & 0.686(3ep) & 0.767(3ep)   \\
        ST& DeiT Backbone & 0.7332 & 0.7659 & 0.3291 & 0.3870 & 0.6922 & 0.7717  \\
        MT2 & DeiT Backbone & 0.7697 & 0.7956 & 0.3635 & 0.4250 & 0.6900 & 0.7641  \\
        ST& MobileVit xSmall Backbone & 0.6376 & 0.6786 & 0.2239 & 0.2739 & 0.5886 & 0.6708  \\
        MT2 & MobileVit xSmall Backbone & 0.7241 & 0.7597 & 0.2984 & 0.3604 & 0.6770 & 0.7501 \\
        MT2 & EfficientFormerl1 Backbone 256 & 0.7568 & 0.7866 & 0.3515(5ep) & 0.409(5ep) & 0.736(3ep) & 0.8075(3ep) \\
        MT2 & EfficientFormerl1 Backbone 448 & 0.7929 & 0.8139 & 0.3840 & 0.4391 & 0.76631(3ep) & 0.8345(3ep)   \\
        ST& EfficientFormerl3 Backbone 256 & 0.7086 & 0.7398 & 0.2977(5ep) & 0.3889(5ep) & 0.6831(3ep) & 0.7582(3ep)  \\
        ST& EfficientFormerl3 Backbone 512 & 0.7642 & 0.7904 & 0.3364 & 0.3892 & 0.7232(3ep) & 0.7938(3ep)  \\
        MT2 & EfficientFormerl3 Backbone 256 & 0.7694(3ep) & 0.8007(3ep) & 0.3651(3ep) & 0.4242(3ep) & 0.7212(3ep) & 0.7987(3ep)  \\
        MT2 & EfficientFormerl3 Backbone 512 & 0.8053 & 0.8291 & \textbf{0.4116} & \textbf{0.4659 }& \textbf{0.78} & 0.84 \\
        \bottomrule
    \end{tabularx}
\end{table*}

While we have only employed triplet learning with an EfficientNetB0 backbone, the triplet embeddings outperform the supervised learning embeddings in the evaluation task closest to the metric learning task they were trained on. Even when strong pretrained backbones are available, there is no substitute for a carefully crafted model and finetuning. To evaluate uncertainty, we ran several iterations of training for both the EfficientNetB0 and the EfficientFormerl3, our final contenders, with different random seeds and the standard error of accuracy estimates is negligible. Table \ref{tab:train_hp_params} gives the hyperparameters for finetuning all models.

\begin{table*}
    \centering
  \caption{Finetuning Hyperparameters. MT stands for ``multitask''.}
  \label{tab:train_hp_params}
  \tiny
  \begin{tabularx}{\linewidth}{Xrrrrrr}
    \toprule
    Config & Triplet EfficientNetB0 & MT EfficientNetB0 & MT ViT B16 & MT EfficientFormerl3 & MT EfficientFormerl1\\
    \midrule 
    Optimizer & Adam & Adam & Adam & AdamW \cite{loshchilov2018decoupled} & AdamW \\
    Adam $\beta_1$ & 0.9 & 0.9 & 0.9 & 0.9 & 0.9 \\
    Adam $\beta_2$ & 0.999 & 0.999 & 0.999 & 0.999 & 0.999 \\
    Adam $\epsilon$ & 1e-08 & 1e-08 & 1e-07 & 1e-07 & 1e-07\\
    Weight Decay & 0.00 & 0.00 & 0.01 & 0.05 & 0.01\\
    Average Step Time & 21ms & 200ms & 76ms & 37ms & 18ms \\
    Batch Size (train) & 512 & 256 & 64 & 128 & 64\\
    Batch Size (validation) & 256 & 256 & 128 & 128 & 128\\
    Input Size & 224x224x3 & 224x224x3 & 224x224x3 & 224x224x3 & 224x224x3 \\
    Machine Type (training) & 2 P100 & 2 P100 & 2 P100 & 2 P100 & 2 P100\\
    Machine Type (validation) & 2 P100 & 2 P100 & 2 P100 & 2 P100 & 2 P100\\
    Learning Rate & 0.0001 & 0.0001 & 0.0001 & 0.0008 & 0.0002\\
    Loss & Triplet & CE & CE & CE & CE\\
    Learning Rate Scheduler & cosine \cite{loshchilov2016sgdr} & cosine & polynomial & cosine & polynomial\\
    Num Epochs & 10 & 10 & 5 & 12 & 10\\
    Num Parameters New Layers (trainable) & 737,792 & 1,260,014 & 719,598 & 653,038 & 637,100\\
    Num Parameters (trainable) & 2,977,472 & 3,888,234 & 87,109,870 & 31,033,550 & 12,029,094\\
    Num Parameters (total) & 4,373,155 & 4,895,889 & 87,111,918 & 31,101,586 & 12,063,415\\
    Triplet Margin & 0.2 & - & - & - & -\\
    Num Train Examples  & 802,822 & 715,440 & 715,440 & 715,440 & 715,440\\
    Num Validation Examples & 91,576 & 78,000 & 78,000 & 78,000 & 78,000\\
  \bottomrule
    \end{tabularx}
\end{table*}

In a first iteration we evaluated the EfficientNetB0 256d and the EfficientFormerl3 512d representations (finetuned in MT2) using the generated text-to-image queries dataset. On this task, the CNN achieves the best Recall@10 at epoch six (0.0438). The efficient ViT continues learning for all 12 epochs and achieves a slightly higher (+2\%) recall. This is a challenging evaluation tasks for a few reasons. The clicks are biased toward the production search-by-text system that produced them so absolute values are less relevant. The generative diffusion model displays higher errors than in controlled experiments when employed with actual user queries, which are short and lacking garment detail. Prompt engineering for text queries may improve the process. This is early research which hopefully opens a path for other generative AI-fueled approaches.

\subsection{Offline Experiment Results in Downstream Tasks }

In Table \ref{tab:downstream_task_offline} we compare how each training architecture performs offline in two downstream tasks: an ad click prediction task, where the baseline is a standard click-through rate prediction (CTR) model with no visual inputs, and a candidate generator for recommendation ads, where the baseline is an embedding model without visual inputs. The inclusion of EfficientNetB0 based triplet learned embedding leads to the largest AUC metrics lift in the CTR model, and in the retrieval task all finetuned variants perform similarly, with EfficientNetB0 multitask outperforming others slightly in Recall@1000. However, we have not evaluated the EfficientFormer-learned representations yet in these downstream settings.

\begin{table}[ht]
    \centering
    \tiny
    \caption{Ablation studies showing offline lift for various pretrained visual representations used in the downstream click prediction task in ad ranking (CTR), and candidate generation for recommendation ads (AIR).}
    \label{tab:downstream_task_offline}
    \begin{tabularx}{\linewidth}{XXrrrr}
        \toprule
        Training Architecture & Model Architecture & \multicolumn{2}{c}{CTR} & \multicolumn{2}{c}{IR} \\ 
                              &                    &  ROC-AUC & PR-AUC & R@100 & R@1000 \\
        \midrule
        No Visual Inputs  & No Visual Inputs & - & - & - & - \\
        Triplet & EfficientNetB0 Backbone & \textbf{+0.64\%} & \textbf{+2.04\% } & \textbf{+7.58\%} & +6.91\% \\
        Multitask & EfficientNetB0 Backbone & +0.56\% & 0.0\% & +5.93\% & \textbf{+6.98\%} \\
        Multitask & ViT Base16 Backbone & +0.62\% & +1.9\% & +4.38\% & +6.05\% \\
        Pseudo Zero-Shot & ViT Base16 Backbone & +0.32\% & +0.47\% & +0.09\% & +0.75\% \\
        \bottomrule
    \end{tabularx}
\end{table}

\subsection{Online Experiment Results in Downstream Tasks A/B Tests}

\begin{table*}[htbp]
    \centering
    \tiny
    \caption{ Statistically significant online A/B experiment results (p-value < 0.001) using visual representations in e-commerce applications}
    \label{tab:online_exp_results}
    \begin{tabularx}{\linewidth}{Xrrrrrr}
        \toprule
        Online Metric & Ad Information Retrieval & Ad Similar Items Recommendations  & \multicolumn{2}{c}{Visually Similar Recommendations} & \multicolumn{2}{c}{Sponsored-Search Ranking} \\
               &   Ad Recommendations Matching  &   Ad Recommendations Ranker    &  iOS  & Android                         &  CTR & PCCVR \\
        \midrule
        Ad-Recommendation Clicked & +3.97\% & +6.25\% & +0.73\% & +0.11\% & - & - \\
        Ad-Sitewide-CTR     & +0.97\% & +0.77\% &	-0.65\% & +0.38\% &	+2.55\% & -0.02\% \\
        Ad-Sitewide-Clicks  & +1.05\% & +0.70\% & +0.73\% & +1.61\% & +2.07\% & -0.48\% \\
        Ad-Return-On-Spend  & -0.40\% & +0.31\% & +1.92\% &	-       & +1.02\% &	+3.51\% \\
        Ads-Post-Click Purchase Rate &    -	  &    -    &	-     &	+1.18\%	& +0.29\% &	+1.26\% \\
        \bottomrule
    \end{tabularx}
\end{table*}

We include online results from the EfficientNetB0 MT2 tested across visual applications on an e-commerce platform with results in Table \ref{tab:online_exp_results}. Additionally, the EfficientFormerl3 MT2 led to a $+0.65\%$ lift in CTR and purchase rate in a first visually similar ads experiment. In Table \ref{tab:online_exp_results} we can see lift in metrics when including visual representations in sponsored search post-click conversion rate models \cite{awad2023adsformers} and recommendation ranking functions, as well as in learning other embeddings for candidates generation \cite{awad2023adsformers}. The EfficientNetB0 MT2 representations first powered visually similar ad recommendations in Figure \ref{fig:visual_apps}, as well as the first search-by-image shopping experience on the same e-commerce platform. Users search using photos taken with their mobile phone’s camera and the query image embedding is inferred efficiently online.

\section{Conclusion}

In this article we present a holistic approach to learning and evaluating image representations in an efficient manner and the benefits derived by employing these representations in downstream tasks, such as visually similar recommendations. We contrast multiple architectures employed as backbones for representation learning, as well as transfer learning methods. We introduce four offline evaluation approaches, including a novel text-to-image generative method. We present ablation studies and offline and online results in multiple downstream tasks. Learning visual representations is of paramount importance in visually rich e-commerce and online fashion applications. Doing so efficiently is a challenging goal made possible by advances in the field of efficient deep learning in computer vision.

\bibliographystyle{ACM-Reference-Format}
\bibliography{bibliography}

\end{document}